\documentclass[10pt,twocolumn,letterpaper]{article}
\usepackage[pagenumbers]{cvpr} 

\usepackage{graphicx}
\usepackage{amsmath}
\usepackage{amssymb}
\usepackage{booktabs}
\usepackage{amsfonts,bm,mathtools,amsthm}

\usepackage[pagebackref,breaklinks,colorlinks]{hyperref}

\usepackage[capitalize]{cleveref}
\usepackage{multirow}
\usepackage{adjustbox}
\crefname{section}{Sec.}{Secs.}
\Crefname{section}{Section}{Sections}
\Crefname{section}{Section}{Sections}
\Crefname{table}{Table}{Tables}
\crefname{table}{Tab.}{Tabs.}

\newcommand{\myparagraph}[1]{
\vspace{3pt}
\noindent\textbf{#1}}

\newcommand{\Ours}{DiffRF\xspace}
\newcommand{\aka}{\emph{a.k.a.}\xspace}

\begin{document}

\title{\vspace{-30pt}DiffRF: Rendering-Guided 3D Radiance Field Diffusion}

\author{
Norman M{\"u}ller$^{1,2}$~~~
Yawar Siddiqui$^{1,2}$~~~ 
Lorenzo Porzi$^2$~~~
Lorenzo Porzi$^2$~~~
Samuel Rota Bul\`{o}$^2$~~~\\
Peter Kontschieder$^2$~~~
Matthias Nie{\ss}ner$^1$~~~
\vspace{0.2cm} \\
Technical University of Munich$^1$~~~
Meta Reality Labs Zurich$^2$
\vspace{0.2cm}
}


\twocolumn[{%
	\renewcommand\twocolumn[1][]{#1}%
	\maketitle
        \vspace{-1.1cm}
	\begin{center}

            \includegraphics[width=\linewidth, trim={0 4cm 7.2cm 0},clip, page=1]{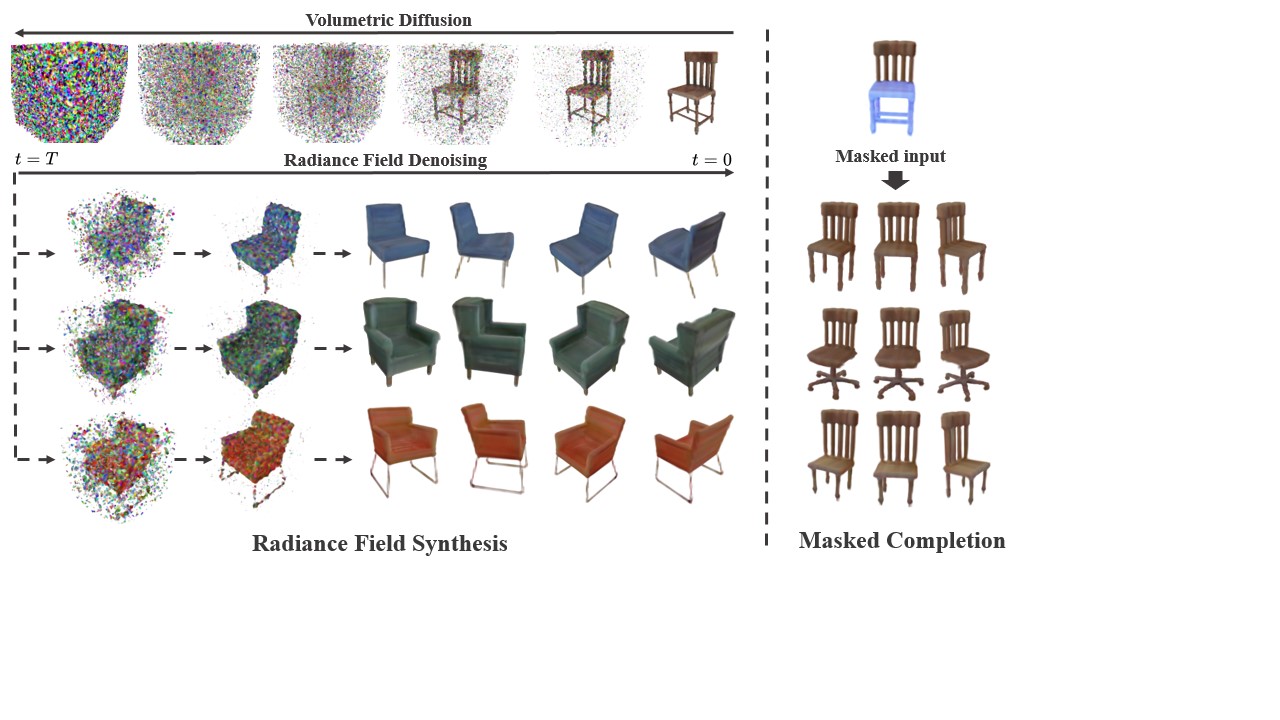}
         \captionof{figure}{
            Our method performs denoising of a probabilistic diffusion process applied to 3D radiance fields. Guided by 3D supervision and volumetric rendering, our model enables the unconditional synthesis of high-fidelity 3D assets (left). We further introduce the novel application of \textit{masked completion} (right), \ie, the task of recovering shape and appearance from incomplete objects (highlighted in light-blue on the top right chair), solved by our model as conditional inference without task-specific training.
            }
		\label{fig:teaser}
	\end{center}    
	\vspace{0.15cm}
}]

\newcommand\blfootnote[1]{%
  \begingroup
  \renewcommand\thefootnote{}\footnote{#1}%
  \addtocounter{footnote}{-1}%
  \endgroup
}

\begin{abstract}
We introduce DiffRF, a novel approach for 3D radiance field synthesis based on denoising diffusion probabilistic models. 
While existing diffusion-based methods operate on images, latent codes, or point cloud data, we are the first to directly generate volumetric radiance fields. 
To this end, we propose a 3D denoising model which directly operates on an explicit voxel grid representation.
However, as radiance fields generated from a set of posed images can be ambiguous and contain artifacts, obtaining ground truth radiance field samples is non-trivial.
We address this challenge by pairing the denoising formulation with a rendering loss, enabling our model to learn a deviated prior that favours good image quality instead of trying to replicate fitting errors like floating artifacts.
In contrast to 2D-diffusion models, our model learns multi-view consistent priors, enabling free-view synthesis and accurate shape generation. 
Compared to 3D GANs, our diffusion-based approach naturally enables conditional generation such as masked completion or single-view 3D synthesis at inference time. 
\blfootnote{Project page: \url{https://sirwyver.github.io/DiffRF/}.}
\end{abstract}
\vspace{-10pt}
\section{Introduction}
\label{sec:intro}
In recent years, Neural Radiance Fields (NeRFs)~\cite{mildenhall2020nerf} have emerged as a powerful representation for fitting individual 3D scenes from posed 2D input images. 
The ability to photo-realistically synthesize novel views from arbitrary viewpoints while respecting the underlying 3D scene geometry has the potential to disrupt and transform applications like AR/VR, gaming, mapping, navigation, \etc. 
A number of recent works have introduced extensions for making NeRFs more sophisticated, by \eg, showing how to incorporate scene semantics~\cite{kundu2022panoptic,fu2022panoptic}, training models from heterogeneous data sources~\cite{martinbrualla2020nerfw}, or scaling them up to represent large-scale scenes~\cite{tancik2022blocknerf,Turki_2022_CVPR}. 
These advances are testament to the versatility of ML-based scene representations; however, they still fit to specific, individual scenes rather than generalizing beyond their input training data.

In contrast, neural field representations that generalize to multiple object categories or learn priors for scenes across datasets appear much more limited to date, despite enabling applications like single-image 3D object generation~\cite{chanmonteiro2020pi-GAN,schwarz2020graf,yu2021pixelnerf,Muller_2022_CVPR,Wang_2021_CVPR} and unconstrained scene exploration~\cite{DeVries_2021_ICCV}. 
These methods explore ways to disentangle object priors into shape and appearance-based components, or to decompose radiance fields into several small and locally-conditioned radiance fields to improve scene generation quality; however, their results still leave significant gaps \wrt photorealism and geometric accuracy.

Directions involving generative adversarial networks (GANs) that have been extended from the 2D domain to 3D-aware neural fields generation are demonstrating impressive synthesis results~\cite{Chan2022}. 
Like regular 2D GANs, the training objective is based on discriminating 2D images, which are obtained by rendering synthesized 3D radiance fields. 

At the same time, diffusion-based models~\cite{sohl2015deep} have recently taken the computer vision research community by storm, performing on-par or even surpassing GANs on multiple 2D benchmarks, and are producing photo-realistic images that are almost indistinguishable from real photographs. 
For multi-modal or conditional settings such as text-to-image synthesis, we currently observe unprecedented output quality and diversity from diffusion-based approaches. 
While several works address purely geometric representations~\cite{zhou20213d,luo2021diffusion}, lifting the denoising-diffusion formulation directly to 3D volumetric radiance fields remains challenging. 
The main reason lies in the nature of diffusion models, which require a one-to-one mapping between the noise vector and the corresponding ground truth data samples.
In the context of radiance fields, such volumetric ground truth data is practically infeasible to obtain, since even running a costly per-sample NeRF optimization results in incomplete and imperfect radiance field reconstructions.

In this work, we present the first diffusion-based generative model that directly synthesizes 3D radiance fields, thus unlocking high-quality 3D asset generation for both shape and appearance. 
Our goal is to learn such a generative model trained across objects, where each sample is given by a set of posed RGB images.

To this end, we propose a 3D denoising model directly operating on an explicit voxel grid representation (Fig.~\ref{fig:teaser}, left) producing high-frequency noise estimates. 
To address the ambiguous and imperfect radiance field representation for each training sample, we propose to bias the noise prediction formulation from Denoising Diffusion Probabilistic Models (DDPMs) towards synthesizing higher image quality by an additional volumetric rendering loss on the estimates. 
This enables our method to learn radiance field priors less prone to fitting artifacts or noise accumulation during the sampling process.
We show that our formulation leads to diverse and geometrically-accurate radiance field synthesis producing efficient, realistic, and view-consistent renderings.
Our learned diffusion prior can be applied in an unconditional setting where 3D object synthesis is obtained in a multi-view consistent way, generating highly-accurate 3D shapes and allowing for free-view synthesis. 
We further introduce the new task of \textit{conditional masked completion} -- analog to shape completion -- for radiance field completion at inference time. 
In this setting, we allow for realistic 3D completion of partially-masked objects without the need for task-specific model adaptation or training (see Fig.~\ref{fig:teaser}, right).

\smallskip\noindent
We summarize our contributions as follows:
\begin{itemize}
\vspace{-0.1cm}
\setlength\itemsep{-.3em}
\item To the best of our knowledge, we introduce the first diffusion model to operate directly on 3D radiance fields, enabling high-quality, truthful 3D geometry and image synthesis.

\item We introduce the novel application of 3D radiance field masked completion, which can be interpreted as a natural extension of image inpainting to the volumetric domain.
\item We show compelling results in unconditional and conditional settings, \eg, by improving over GAN-based approaches on image quality (from 16.54 to 15.95 in FID) and geometry synthesis (improving MMD from 5.62 to 4.42), on the challenging PhotoShape Chairs dataset~\cite{photoshape2018}. 
\end{itemize}

\section{Related work}

\myparagraph{Diffusion models.}
Since the seminal work by Sohl-Dickstein~\etal~\cite{sohl2015deep} on generative diffusion modeling, two classes of generative models have been proposed that perform inversion of a diffusion process: Denoising Score Matching (DSM)~\cite{song2020denoising,song2020improved,genmodel} and Denoising Diffusion Probablistic Models (DDPMs)~\cite{ho2020denoising}.
Both approaches have shown to be flavours of a single framework, coined Score SDE, in the work of Song~\etal~\cite{song2020score}.
The term ``Diffusion models'' is now being used as an all-encompassing name for this constellation of methods.
Several main directions being studied include devising different sampling schemes~\cite{Karras2022edm,song2021denoising} and noising models~\cite{daras2022soft,hoogeboom2022blurring}, exploring alternative formulations and training algorithms~\cite{song2021maximum,huang2021variational,nichol2021improved}, and improving efficiency~\cite{rombach2021highresolution}.
An overview of current research results is given by Karras~\etal~\cite{Karras2022edm}.

Diffusion models have been used to obtain state-of-the-art results in many domains such as text-to-image and guided synthesis \cite{GLIDE,rombach2021highresolution,meng2022sdedit,poole2022dreamfusion}, 3D shape generation \cite{zhou20213d,luo2021diffusion,gaishape, zeng2022lion}, molecule prediction \cite{trippe2022diffusion,xu2022geodiff,luo2021predicting}, and video generation \cite{ho2022video,yang2022diffusion}.
Interestingly, diffusion models have shown to outperform Generative Adversarial Networks (GANs) in high-resolution image generation tasks~\cite{dhariwal2021diffusion,rombach2021highresolution}, achieving unprecedented results in conditional image generation~\cite{ramesh2022hierarchical}.
Furthermore, compared to GANs, which are often prone to divergence and mode collapse~\cite{brock2018large,miyato2018spectral}, diffusion models have been observed to be much easier to train, although training time is still relatively long.

\myparagraph{3D generation.}
Initially developed for 2D image synthesis, adversarial approaches have also found success in 3D, \eg, for generating meshes~\cite{wu2016learning,gao2022get3d}, 3D textures~\cite{siddiqui2022texturify}, voxelized representations~\cite{chen2018text2shape,henzler2019escaping}, or Neural Radiance Fields (NeRFs)~\cite{chanmonteiro2020pi-GAN,schwarz2020graf,niemeyer2021giraffe,Chan2022,gu2021stylenerf,zhou2021cips}.
In particular, methods in this last category have received much attention in recent years, as they can be trained purely from collections of 2D images, without any form of 3D supervision, and enable for the first time photo-realistic novel view synthesis of the generated 3D objects.
Pi-GAN~\cite{chanmonteiro2020pi-GAN} and GRAF~\cite{schwarz2020graf} propose similar approaches where a standard NeRF~\cite{mildenhall2020nerf} model is cast in a GAN setting by adding a form of stochastic conditioning, trained with an adversarial loss.
These approaches are partly limited by the high training-time memory cost of NeRF-style volumetric rendering, forcing them to use low-resolution image patches.
CIPS-3D~\cite{zhou2021cips} and GIRAFFE~\cite{niemeyer2021giraffe} solve this issue by letting the volume rendering component output a low-resolution 2D feature map, which is then upsampled by an efficient convolutional network to produce the final image.
This approach drastically improves the quality and resolution of the rendered images, but also introduces 3D inconsistencies, as the convolutional stage can process different views of the same object in arbitrarily different ways.
StyleNeRF~\cite{gu2021stylenerf} partially addresses this problem by carefully designing the convolutional stage to minimize inconsistencies, while EG3D~\cite{Chan2022} further improves on training efficiency by replacing the MLP-based NeRF with a light-weight tri-plane volumetric model generated by a convolutional network.
In contrast to our method, these GAN-based approaches do not naturally support conditional synthesis or completion.

Compared to GANs, diffusion models are relatively under-explored as a tool for 3D synthesis, but a few works have emerged in the past two years.
Some diffusion based-generators have been proposed for 3D point clouds~\cite{zhou20213d,zeng2022lion,luo2021diffusion,gaishape},  showing promising results for conditional synthesis, completion, and other related tasks~\cite{zeng2022lion,zhou20213d}.
DreamFusion~\cite{poole2022dreamfusion}, 3DDesigner~\cite{li20223ddesigner},  and GAUDI~\cite{bautista2022gaudi}, like our work, employ diffusion models in conjunction with radiance fields, with applications in both conditional (on text and images) and unconditional 3D generation.
DreamFusion~\cite{poole2022dreamfusion} presents an algorithm to generate NeRFs, augmented with an illumination component, by optimizing a loss defined by a pre-trained 2D text-conditional diffusion model.
GAUDI~\cite{bautista2022gaudi} builds a 3D scene generator by first training a conditional NeRF to reconstruct a set of indoor videos given scene-specific latents, and then fitting a diffusion model to capture the learned latent space.
Concurrent works like~\cite{anciukevicius2022renderdiffusion,wang2022rodin} apply the denoising-diffusion approach to factorized radiance representations.
In contrast, our diffusion model operates directly in the space of radiance fields which directly enables 3D-conditional tasks like shape completion (\cref{sec:conditional-generation}) by leveraging the learned volumetric prior.

\newcommand{\rvF}{{\mathtt F}}

\section{Method}
Our method consists of a generative model for 3D objects that builds on recent state-of-the-art diffusion probabilistic models~\cite{ho2020denoising}. It is trained to revert a process that gradually corrupts 3D objects by injecting noise at different scales. In our case, 3D objects are represented as radiance fields~\cite{mildenhall2020nerf}, so the learned denoising process allows our method to generate object radiance fields from noise.

\begin{figure*}
\begin{center}
\includegraphics[width=.98\textwidth, trim={0 10.3cm 0 0},clip]{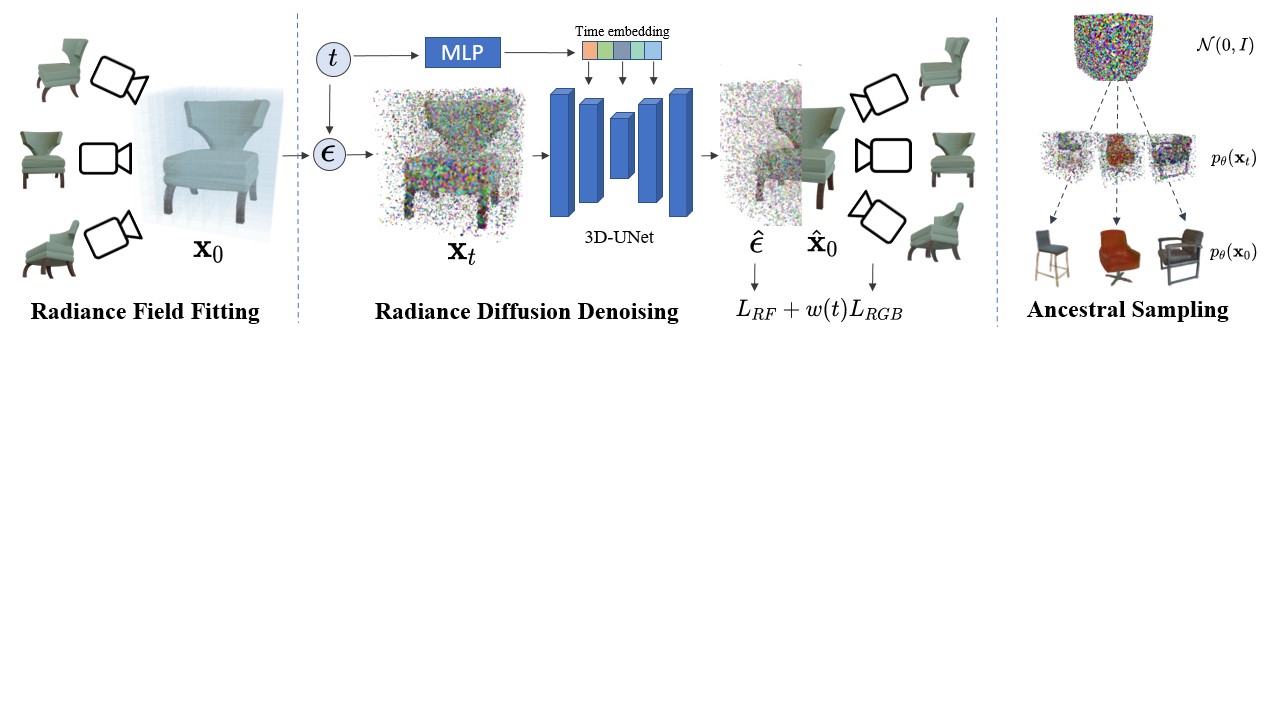}

\end{center}
\vspace{-1em}
\caption{
For a time step $t$ uniformly sampled from ${1,...,T}$, we first diffuse an initial radiance field $f_0$ according to a fixed noising schedule. The resulting $f_t$ is passed through a time-conditioned 3D-UNet, giving an estimate of the applied noise $\epsilon$. We guide the model by the noise prediction loss $L_\mathtt{RF}$ as well as a rendering loss $L_\mathtt{RGB}$ on the  predicted denoising $\tilde{f}_0$. 
}
\label{fig:method}
\end{figure*}

Since we target the generation of 3D objects as radiance fields, we begin with a brief overview of this representation before delving into the details of our method.

\subsection{Radiance Fields} A radiance field $(\sigma, \xi)$ is an implicit, volumetric representation of a 3D object that is given in terms of a density field $\sigma:\mathcal X\to\mathbb R_+$ and an RGB color field $\xi:\mathcal X\to\mathbb R^3$ defined over a 3D domain $\mathcal X\subset\mathbb R^3$.\footnote{Typically the color field spans also the viewing direction, but we omit it in this work.} The density field gives information about the presence of an object in a specific point in space, whereas the color field provides the eventual corresponding RGB color.

Following a ray casting logic, the radiance field can be rendered along a given ray $r$ yielding an RGB color $c_r$
with the following equation~\cite{mildenhall2020nerf}
\begin{equation}
\label{eq:rendering}
    c_r\coloneqq \int_0^\infty\tau_s(r)\sigma(r_s)\xi(r_s)ds \,,
\end{equation}
where a ray $r$ is a linear curve parametrized by $s$ with unit velocity, $r_s\in\mathcal X$ denotes the point along the ray at $s$, and $\tau_s(r)$ is the transmittance probability at $s$, which is given by
\begin{equation}
    \mathrm \tau_s(r)\coloneqq\exp\left(-\int_0^s\sigma(s')ds'\right)\,.
\end{equation}
Using the ray rendering equation above, we can render the radiance field from any given camera, yielding an image of the novel view. It is sufficient to turn the camera into a proper set of rays expressed in world coordinates.

There exist various ways of implementing a radiance field, ranging from neural networks~\cite{mildenhall2020nerf} to explicit voxel grids~\cite{ReluField_sigg_22,SunSC22}. In this work, we opt for the latter, since it enables good rendering quality along with faster training and inference. The explicit grid can be queried at continuous positions via bilinear interpolation of the voxel vertices. 

Under the explicit representation, radiance fields become 4D tensors, where the first three dimensions index a grid spanning $\mathcal X$, whereas the last dimension indexes the density and color channels.

\subsection{Generating Radiance Fields}
Following recent advancements in the context of denosing-based generative methods~\cite{ho2020denoising}, we formulate our generative model of radiance fields as a denoising diffusion probabilistic model. 

\myparagraph{Generation process.}
The generation (\aka~denoising) process is governed by a discrete-time Markov chain defined on the state space $\mathcal F$ of all possible pre-activated radiance fields expressed as flattened 4D tensors of fixed size.\footnote{We consider pre-activated radiance fields, where both density and RGB color channels span a linear space and we assume a proper activation function will be applied at the time of rendering. This is required to have additive noise, while preserving a valid radiance field representation.}
The chain has a finite number of time steps $\{0,\ldots,T\}$.
The denoising process starts by sampling a state $f_T$ from a standard, multivariate normal distribution $p(f_T)\coloneqq\mathcal N(f_T|0,I)$, and generates states $f_{t-1}$ from $f_t$ by leveraging reversed transition probabilities $p_\theta(f_{t-1}|f_t)$ that are Gaussian with learned parameters $\theta$. Specifically we have
\begin{equation}
\label{eq:bwd_transition}
    p_\theta(f_{t-1}|f_t)\coloneqq \mathcal N(f_{t-1}|\mu_\theta(f_t,t),\Sigma_t)\,.
\end{equation}
The generation process iterates up to the final state $f_0$, which represents the radiance field of a 3D object generated by our method.
The mean of the Gaussian in \eqref{eq:bwd_transition} can be directly modeled with a neural network. However, as we will see later, it is more convenient to consider the following reparametrization of it
\begin{equation} 
\label{eq:mu}
    \mu_\theta(f_t,t)\coloneqq a_t (f_t-b_t\epsilon_\theta(f_t,t))\,,
\end{equation}
where $\epsilon_\theta(f_t,t)$ is the noise that has been used to corrupt $f_{t-1}$ predicted by \eg a neural network, whereas $a_t$ and $b_t$ are pre-defined coefficients.
Also the covariance $\Sigma_t$ takes a pre-defined value, although it could be data-dependent. Additional details about the value that the pre-defined variables take are given in \cref{ss:training}.

\myparagraph{Diffusion process.}
While the generation process works by iteratively denoising a completely random radiance field, the diffusion process works the other way around and iteratively corrupts samples from the distribution of 3D objects we want to model. We introduce it because it plays a fundamental role in the training scheme of the generation process. The diffusion process is governed by a discrete-time Markov chain with the same state space and time bounds mentioned in the generation process but with Gaussian transition probabilities that are pre-defined and given by
\begin{equation}
q(f_t|f_{t-1})\coloneqq\mathcal N(f_t|\sqrt{\alpha_t}f_{t-1},\beta_t I)\,,
\end{equation}
where $\alpha_t\coloneqq 1-\beta_t$ and $0\leq\beta_t\leq 1$ are predefined coefficients implementing a schedule for the injected noise variance.
The process starts by picking $f_0$ from the distribution $q(f_0)$ of 3D object radiance fields we want to model, iteratively samples $f_t$ given $f_{t-1}$ yielding a scaled and noise-corrupted version of the latter,
and stops with $f_T$ being typically close to completely random depending on the implemented noise-variance schedule.
By exploiting properties of the Gaussian distribution, we can conveniently express the distribution of $f_t$ conditioned on $f_0$ directly as a Gaussian distribution, yielding
\begin{equation}
\label{eq:direct_fwd}
q(f_t|f_0)=\mathcal N(f_t|\sqrt{\bar\alpha_t}f_0,(1-\bar\alpha_t) I)\,,
\end{equation}
where $\bar\alpha_t\coloneqq\prod_{i=1}^t\alpha_i$. This relation will be useful to quickly generate diffused data points at arbitrary time steps.

\subsection{Training Objective}\label{ss:training}
Our training objective comprises two complementary losses: i) A loss $L_\mathtt{RF}$ that penalizes the generation of radiance fields that do not fit the data distribution, and ii) an RGB loss $L_\mathtt{RGB}$ geared towards improving the quality of renderings from generated radiance fields.

\myparagraph{Radiance field generation loss.}
Following~\cite{ho2020denoising}, we derive the training objective for our model starting from a variational upper-bound on the Negative Log-Likelihood (NLL). This upper-bound requires specifying a surrogate distribution that we refer to as $q$ because it indeed corresponds to the distribution $q$ governing the diffusion process, establishing the anticipated fundamental link with the generation process. We provide here some key steps of the derivation of the bound and refer to~\cite{ho2020denoising} for more details about the intermediate ones. By Jensen inequality, the NLL of a data point $f_0\in\mathcal F$ can be upper-bounded by leveraging $q$ as follows:
\begin{equation}
    -\log p_\theta(f_0)\leq\mathbb E_q\left[-\log\frac{p_\theta(f_{0:T})}{q(f_{1:T}|f_0)}\right]\coloneqq L_\mathtt{RF}(f_0|\theta)\,,
\end{equation}
where $f_{t_1:t_2}$ stands for $(f_{t_1},\ldots,f_{t_2})$. 
The loss $L_\mathtt{RF}(f_0|\theta)$ bounding the NLL can be further decomposed into the following sum, up to a constant independent from $\theta$
\begin{equation}
\label{eq:Lk}
    L_\mathtt{RF}(f_0|\theta)= \sum_{t=1}^{T} L_\mathtt{RF}^{t}(f_0|\theta)+\text{const}\,.
\end{equation}
Here, $L_\mathtt{RF}^{t}(f_0|\theta)$ takes a simple and intuitive form if we set $a_t\coloneqq\frac{1}{\sqrt{\alpha_t}}$ and $b_t\coloneqq\frac{\beta_t}{\sqrt{1-\bar\alpha_t}}$ in \eqref{eq:mu}, and pick $\Sigma_t\coloneqq\frac{\beta_t^2}{2\alpha_t(1-\bar\alpha_t)}I$. Indeed, it yields 
\begin{align*}
    L^{t}_\mathtt{RF}(f_0|\theta)&\coloneqq\mathbb E_q\left[\left\Vert\epsilon-\epsilon_\theta(f_{t},t)\right\Vert^2\right]\\
    &=\mathbb E_\phi\left[\left\Vert\epsilon-\epsilon_\theta(\sqrt{\bar\alpha_{t}}f_0+\sqrt{1-\bar\alpha_{t}}\epsilon,t)\right\Vert^2\right]\,,
\end{align*}
where $\phi(\epsilon)\coloneqq\mathcal N(\epsilon|0,I)$ is the probability distribution of $\epsilon$, namely a normal multivariate, and the last equality follows from \eqref{eq:direct_fwd}.

\myparagraph{Radiance field rendering loss.}
We complement the previous loss with an additional RGB loss $L_\mathtt{RGB}(f_0|\theta)$, aimed at improving the quality of renderings from generated radiance fields. Indeed, the Euclidean metric on the representation that is implicitly used in the previous loss to assess the quality of generated radiance fields does not necessarily ensure the absence of artifacts once we try to render the radiance field.
We define $L_\mathtt{RGB}(f_0|\theta)$ as a sum of time-specific terms $L^t_\mathtt{RGB}(f_0|\theta)$ similar to~\eqref{eq:Lk}, yielding
\begin{equation}
    L_\mathtt{RGB}(f_0|\theta)\coloneqq\sum_{t=1}^T L^t_\mathtt{RGB}(f_0|\theta)\,.
\end{equation}
Given a radiance field $f\in\mathcal F$, a viewpoint $v$, we denote by $R(v, f)$ the image obtained after rendering $f$ from viewpoint $v$ using equation~\eqref{eq:rendering}. We also denote by $\ell_v(f,I)$ the Euclidean distance between the rendered images from viewpoint $v$ using radiance fields $f$ and the ground-truth image $I_v$ from the view-point $v$, \ie
\begin{equation}
    \ell_v(f,I)\coloneqq\Vert I_v-R(v,f)\Vert^2\,.
\end{equation}

The idea is to compare the rendering of a given radiance field $f_0$ sampled from the data distribution corrupted with $t$ diffusion steps and then fully denoised against the original ground-truth image $I_v$ used to obtain $f_0$. In theory, this implies sampling first $f_t$ from $q(f_t|f_0)$ and then sampling back $f_0$ from $p_\theta(f_0|f_t)$. This is, however, computationally demanding and we resort to a simpler approximation. From the definition of $L_\mathtt{RF}^t$, the loss pushes towards having $\epsilon\approx\epsilon_\theta(f_t,t)$ from which we can derive the approximation $\tilde f^t_0(\epsilon,\theta)\coloneqq f_0+\frac{\sqrt{1-\bar\alpha_t}}{\sqrt{\bar\alpha_t}}(\epsilon -\epsilon_\theta(f_t,t))$. We can then define our rendering loss as 
\begin{equation}
\label{eq:lk_rgb}
    L^t_\mathtt{RGB}(f_0|\theta)\coloneqq\omega_t\mathbb E_{\phi,\psi}\left[\ell_v(\tilde f^t_0(\epsilon,\theta),I)\right]\,,
\end{equation}
where the expectation is taken with respect to a prior distribution $\psi$ for the viewpoint $v$ and $\epsilon\sim\phi(\epsilon)$. Since the approximation is reasonable only with steps $t$ close to zero, we introduce a weight $w_t$ that decays as step values increase (\eg we use $\omega_t\coloneqq\bar\alpha_t^2$). We provide evidence in the experimental section that despite being an approximation, the proposed loss contributes to significantly improving the results.

\myparagraph{Final loss.} To summarize, the final training loss per data point $f_0$ is given by the weighted combination of the radiance field generation and rendering losses introduced before, with a small variation that enables stochastic sampling of the step $t$ from a uniform distribution $\kappa(t)$:
\begin{align*}
    L(\theta)&\coloneqq L_\mathtt{RF}(f_0|\theta)+\lambda_\mathtt{RGB}L_\mathtt{RGB}(f_0|\theta)\\
    &\propto\mathbb E_{\kappa}\left[L_\mathtt{RF}^t(f_0|\theta)+\lambda_\mathtt{RGB}L_\mathtt{RGB}^t(f_0|\theta)\right].
\end{align*}

\myparagraph{Implementation details}
We implement $\epsilon_\theta$ as a 3D-UNet, which is based on the 2D-UNet architecture introduced in~\cite{dhariwal2021diffusion} by replacing 2D convolutions and attention layers with corresponding 3D operators. 
For training, we uniformly sample timesteps $t=1,\ldots,T=1000$ for all experiments with variances of the diffusion process linearly increasing from $\beta_1 =  0.0015$ to $\beta_T = 0.05$, and choose to weight the rendering loss $L_\mathtt{RGB}^t$ with $w_t=\bar\alpha_t^2$.
We refer to the supplementary material for additional details.

\section{Experiments}\label{sec:exps}
In this section, we evaluate the performance of our method on both unconditional and conditional radiance field generation. 

\myparagraph{Datasets.}
We run experiments on the PhotoShape Chairs~\cite{photoshape2018} and on the Amazon Berkeley Objects (ABO) Tables dataset~\cite{collins2022abo}. For PhotoShape Chairs, we render the provided 15,576 chairs using Blender Cycles \cite{Blender} from 200 views on an Archimedean spiral. For ABO Tables, we use the provided 91 renderings with 2-3 different environment map settings per object, resulting in 1676 tables. 
Since both datasets do not provide radiance field representations of 3D objects, we generate them using a voxel-based approach at a resolution of $32^3$ from the multi-view renderings. 

\myparagraph{Metrics.}
We evaluate image quality using the Fr\'{e}chet Inception Distance~\cite{FID} (FID) and  Kernel Inception Distance~\cite{KID} (KID) using~\cite{obukhov2020torchfidelity}. For comparison of the geometrical quality, we follow~\cite{achlioptas2018learning} and compute the Coverage Score (COV) and Minimum Matching Distance (MMD) using Chamfer Distance (CD). 
While the Coverage Score measures the diversity of the generated samples, MMD assesses the quality of the generated samples. All metrics are evaluated at a resolution of $128\times128$. 

\subsection{Unconditional Radiance Field Synthesis}

\myparagraph{Comparison against state of the art.} 
We quantitatively evaluate our approach on the task of unconditional 3D synthesis on PhotoShape\cite{photoshape2018} in \cref{tab:photoshape} and ABO Tables\cite{collins2022abo} in \cref{tab:abo}. We compare against leading methods for 3D-aware image synthesis:~$\pi$-GAN~\cite{chanmonteiro2020pi-GAN} and EG3D~\cite{Chan2022}. 
Both our method and GAN-based approaches use the same set of rendered images for training. While we pre-process the rendered images to create a radiance field representation for each of the shape samples, the GAN-based methods are trained directly on the rendered images.

\begin{figure*}
\begin{center}
\includegraphics[width=.95\textwidth, trim={0.35cm 4cm 4.4cm 0},clip]{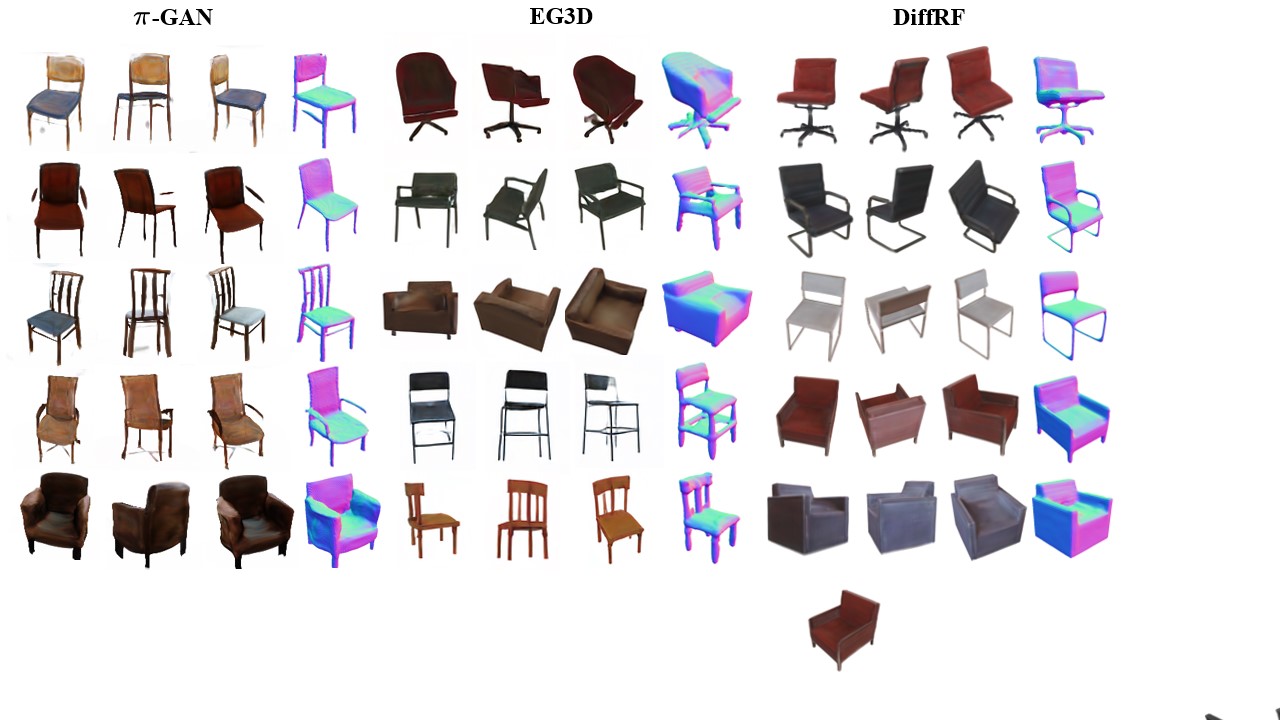}
\end{center}
\vspace{-1em}
\caption{
Qualitative comparison between ~$\pi$-GAN~\cite{chanmonteiro2020pi-GAN}, EG3D~\cite{Chan2022}, and our method on PhotoShape Chairs~\cite{photoshape2018}. Our approach leads to diverse, geometrically accurate models that allow for high-quality renderings.
}
\label{fig:photoshape}
\end{figure*}

\begin{figure*}
\begin{center}
\includegraphics[width=.95\textwidth, trim={0.2cm 5.9cm 0.1cm 0},clip]{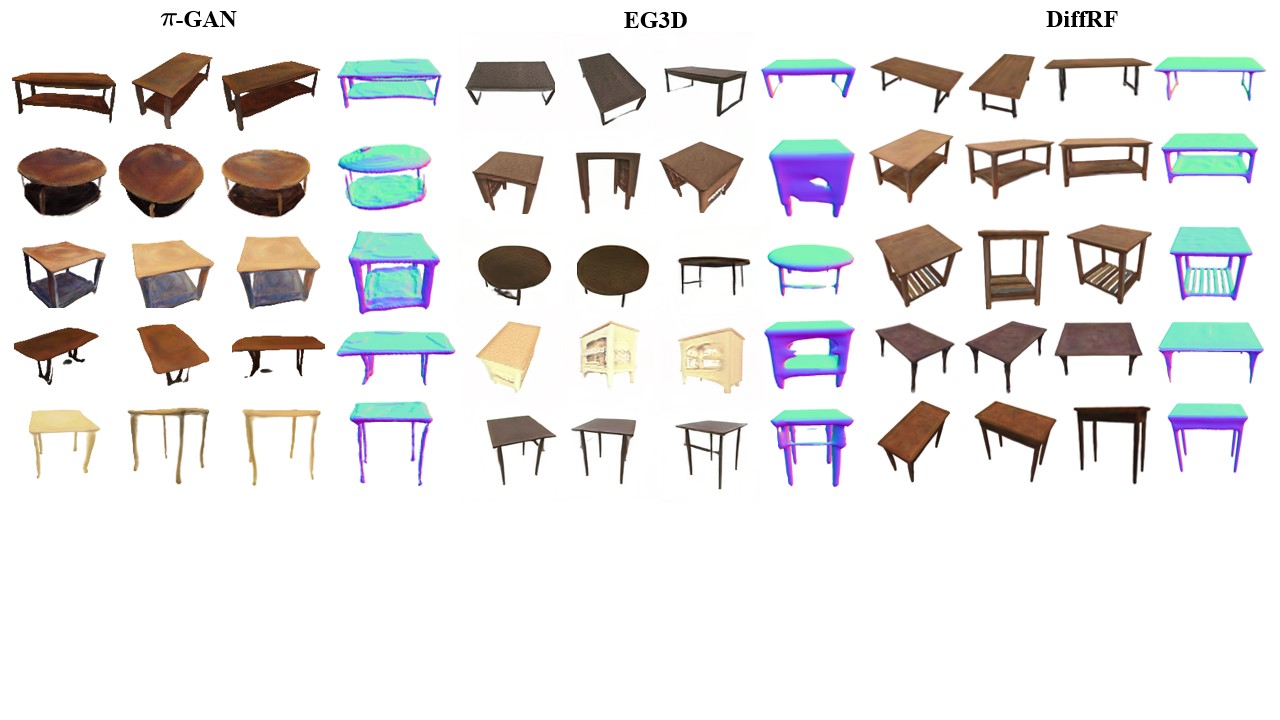}
\end{center}
\vspace{-1em}
\caption{Qualitative comparison between ~$\pi$-GAN~\cite{chanmonteiro2020pi-GAN}, EG3D~\cite{Chan2022}, and our method on ABO Tables~\cite{collins2022abo}. Our approach generates high quality and diverse samples with accurate geometry.}
\label{fig:abo}
\vspace{-5pt}
\end{figure*}

Compared to these approaches, our method yields overall better image quality while achieving significant improvements in geometrical quality and diversity. 
\cref{fig:photoshape} and \cref{fig:abo} show a qualitative comparison of our method with $\pi$-GAN and EG3D. While EG3D achieves good image quality, it tends to produce inaccurate shapes and view-dependent image artifacts, like adding or removing armrests or changing the supporting structure. 
As the training objective (see \cref{ss:training}) is to invert the diffusion process to denoise towards detailed volumetric representations, we observe that \Ours{} reliably generates radiance fields with fine photo-metric and geometrical details.

\myparagraph{Contribution of the rendering loss.} 
\cref{tab:photoshape} and \cref{tab:abo} show ablation results where we evaluate the effects of 2D supervision on the radiance synthesis.
Removing 2D supervision (``\Ours{} w/o 2D'', line 3 in the tables), as expected, has a noticeable effect on FID, which increases by $\approx2.3$ for PhotoShape and by $\approx8.8$ for ABO Tables. This shows that biasing the noise prediction formulation from DDPM by a volumetric rendering loss leads to higher image quality. Qualitative comparisons can be found in the appendix.
We notice a decrease in the Coverage Score that we explain by the fact that the rendering loss guides the denoising model towards learning radiance fields with less artifacts, thus reducing the amount of diverse but spurious shapes.

\begin{table}[t]
\centering
\resizebox{.95\columnwidth}{!}{
\begin{tabular}{lccccc}
\toprule
Method  & FID ↓ & KID ↓  & COV $\uparrow$  & MMD ↓ \\ 
\midrule
$\pi$-GAN~\cite{chanmonteiro2020pi-GAN}   & 52.71    & 13.64  &  39.92    & 7.387     \\
EG3D~\cite{Chan2022}        & 16.54 & 8.412  &  47.55   &  5.619    \\ 
\midrule
\Ours{} w/o 2D &    18.27 &  9.263   & \textbf{59.20 }  & 4.543        \\ 
\Ours{}        & \textbf{15.95} & \textbf{7.935}   &     58.93  &       \textbf{4.416}  \\
\bottomrule
\end{tabular}
}
\captionof{table}{Quantitative comparison of unconditional generation on the PhotoShape Chairs~\cite{photoshape2018} dataset. Our method achieves a better image and geometric quality than state-of-the-art GAN-based approaches. The additional 2D rendering loss improves image quality as indicated by the drop in quality without it. MMD and KID scores are multiplied by $10^3$.}
\label{tab:photoshape}
\end{table}

\begin{table}[t]
\centering
\resizebox{.95\columnwidth}{!}{
\begin{tabular}{lcccc}
\toprule
Method  & FID ↓  & KID ↓ &   COV $\uparrow$  & MMD ↓ \\ 
\midrule
$\pi$-GAN~\cite{chanmonteiro2020pi-GAN} & 41.67   & 13.81 & 44.23  & 10.92     \\
EG3D~\cite{Chan2022}        & 31.18  & 11.67 &  48.15  &  9.327    \\ 
\midrule
\Ours{} w/o 2D  & 35.89 &  13.94   &  \textbf{63.46}     &  8.013    \\ 
\Ours{}        & \textbf{27.06} &   \textbf{10.03}   & 61.54  & \textbf{7.610}  \\ 
\bottomrule
\end{tabular}
}
\captionof{table}{Quantitative comparison of unconditional generation on the ABO Tables~\cite{collins2022abo} dataset. Our method achieves a better image and geometric quality than state-of-the-art GAN-based approaches, with the 2D rendering loss being important. MMD and KID scores are multiplied by $10^3$.}
\label{tab:abo}
\end{table}

\begin{table}[t]
\centering
\resizebox{.95\columnwidth}{!}{
\begin{tabular}{llccccc}
\toprule
\multicolumn{2}{r}{Masking:} & 20\% & 40\% & 60\% & 80\% & Avg \\
\midrule
\multirow{2}{*}{\rotatebox[origin=c]{90}{\footnotesize mPSNR$\uparrow$}}
& EG3D & 23.71 & 24.86 & 24.92 & 25.79 & 24.82 \\[3pt]
& \Ours{} & 24.85 & 26.66 & 28.23 & 30.38 & \textbf{27.53} \\[3pt]
\midrule
\multirow{2}{*}{\rotatebox[origin=c]{90}{\footnotesize FID$\downarrow$}}
& EG3D & 25.91 & 29.41 & 33.06 & 34.31 & 30.67 \\[3pt]
& \Ours{} & 22.36 & 27.74 & 31.16 & 29.84 & \textbf{27.78} \\[3pt]
\bottomrule
\end{tabular}
}
\captionof{table}{Quantitative evaluation on the task of radiance field completion at different levels of masking. EG3D struggles to faithfully reconstruct non-masked regions of the original sample, while our method maintains the non-masked regions and synthesises coherent completions for the masked regions.}
\label{tab:masking_res}
\vspace{-13pt}
\end{table}

\subsection{Conditional Generation}
\label{sec:conditional-generation}
GANs need to be trained in order to be conditioned on a particular task, while diffusion models can be effectively conditioned at test-time~\cite{dhariwal2021diffusion, song2020score, Lugmayr_2022_CVPR}. We leverage this property for the novel task of masked radiance field completion. 

\begin{figure}
\begin{center}
\includegraphics[width=.99\columnwidth, trim={0 13.2cm 23.5cm 0},clip, page=2]{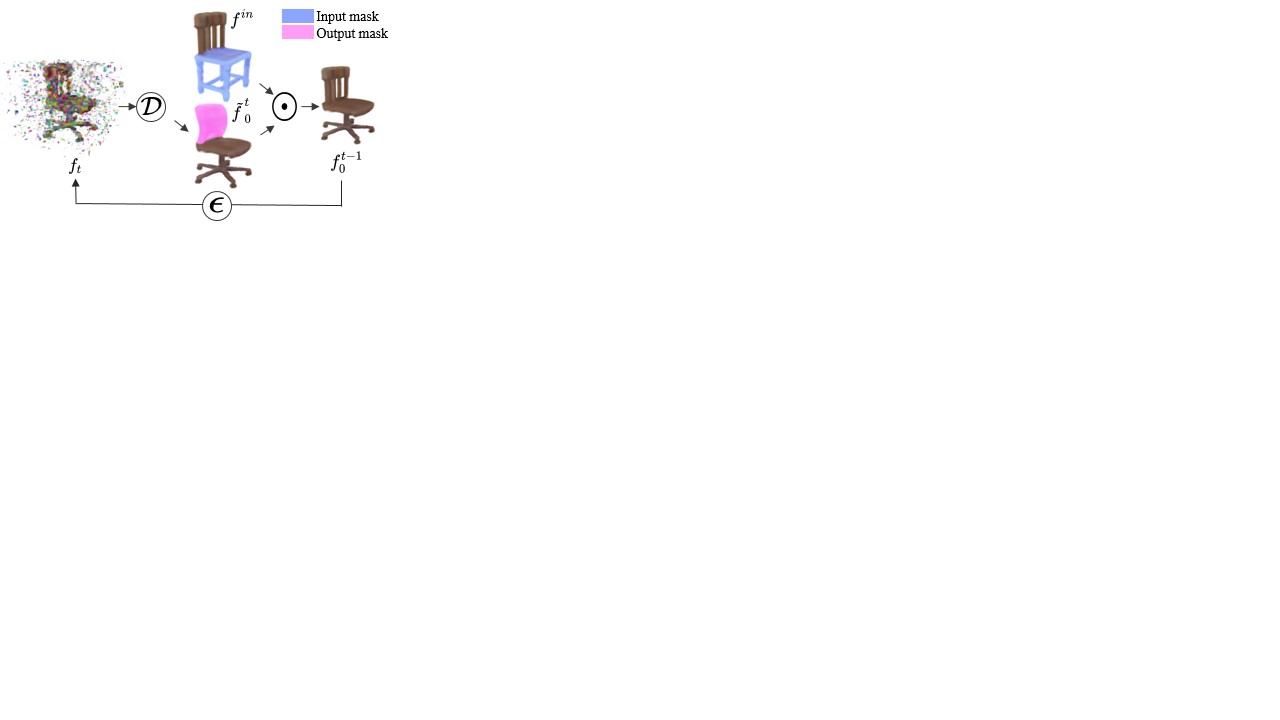}
\end{center}
\vspace{-1em}
\caption{
Masked radiance field completion. For a trained denoiser $\mathcal{D}$ and a given a radiance field with a 3D masked region (shown in blue), completion can be obtained by iteratively fusing the non-masked input region with the estimated denoisings $\tilde{f}^t_0$ and applying the diffusion process on the resulting $f_0^{t-1}$. 
}
\label{fig:masked_completion}
\vspace{-7pt}
\end{figure}

\myparagraph{Masked Radiance Field Completion.}
Shape completion and image inpainting are well-studied tasks ~\cite{pointflow, zhou20213d, yu2018generative,suvorov2021resolution}  aiming to fill missing regions within a geometrical representation or in an image, respectively. 
We propose to combine both in the novel task of masked radiance field completion: Given a radiance field and a 3D mask, synthesize a completion of the masked region that harmonizes with the non-masked region.
Inspired by RePaint~\cite{Lugmayr_2022_CVPR}, we perform conditional completion by gradually guiding the unconditional sampling process in the known region to the input $f^{in}$ 
\begin{align*}
    f_0^{t-1} &= \sqrt{\bar\alpha_t} (m \odot \tilde{f}^t_0 + (1-m) \odot f^{in}) \\
    f_{t-1}& \sim \mathcal N( f_0^{t-1} , (1-\bar\alpha_t) I)\,,
\end{align*}
where $m$ is a binary mask applied to the input (light blue in \cref{fig:masked_completion}) and $\odot$ denotes element-wise multiplication on the voxel grid.

\begin{figure}
\begin{center}
\includegraphics[width=.99\columnwidth, trim={0cm 8cm 17cm 0},clip]{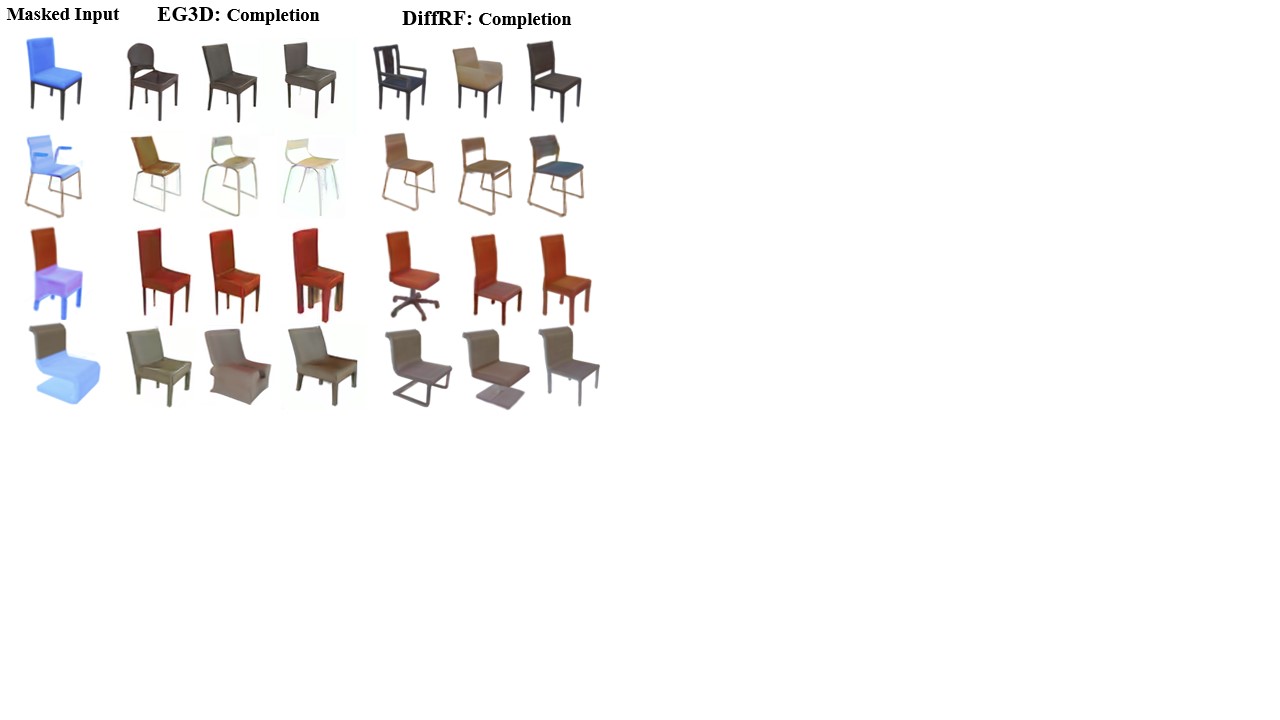}
\end{center}
\vspace{-1em}
\caption{
Qualitative completion of masked chairs from PhotoShape~\cite{photoshape2018}. \Ours{} shows more diverse proposals compared to EG3D, while also maintaining the original non-masked regions. 
}
\label{fig:result_masked_completion}
\vspace{-5pt}
\end{figure}

A quantitative analysis of the masking performance is shown in \cref{tab:masking_res}, where we compare our method against EG3D at different levels of masking. At each level, we randomly mask 200 samples and evaluate the completion performance in terms of FID (by rendering from 10 random views), as well as the photo-metric accuracy of unmasked regions (mPSNR).
For EG3D, we use masked GAN inversion, where we perform global latent optimization (GLO) to minimize the photo-metric error on the re-projected, unmasked regions.
We notice that, due to the single latent code representation, EG3D struggles to faithfully reconstruct the unmasked region of the input sample, and regularization is needed to not corrupt the overall representation. \cref{fig:result_masked_completion} further shows a qualitative comparison of the masking performance against EG3D.

\begin{figure}
\begin{center}
\includegraphics[width=.99\columnwidth, trim={0 3.2cm 12cm 0},clip]{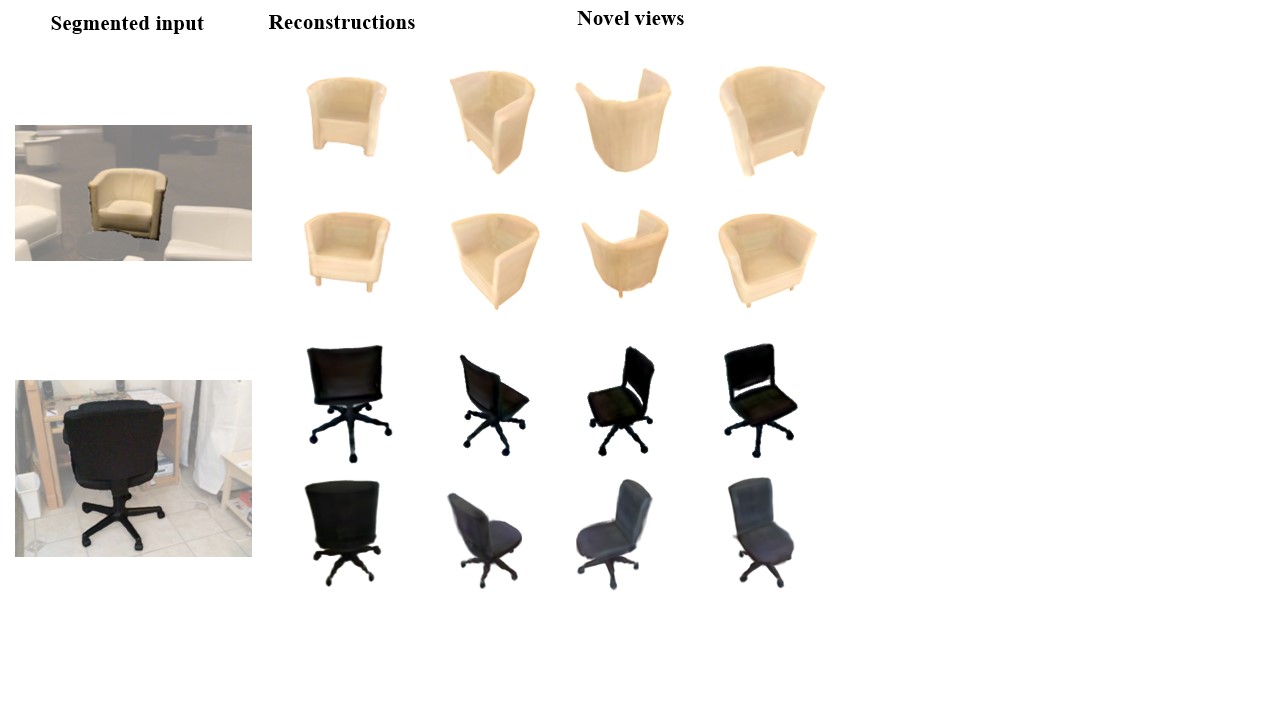}
\end{center}
\vspace{-1em}
\caption{
Qualitative synthesis results of single-view reconstruction on ScanNet~\cite{dai2017scannet}. Given a single posed image and its foreground mask, our method can recover meaningful proposals for the radiance fields describing the objects (shown in separate rows). 
}
\label{fig:scannet_rec}
\vspace{-5pt}
\end{figure}

\myparagraph{Image-to-Volume Synthesis.}
\Ours{} can be used to obtain 3D radiance fields from single view images by steering the sampling process using volumetric rendering. For this, we adopt the Classifer Guidance formulation from~\cite{dhariwal2021diffusion} to guide the denoising process towards minimizing the rendering error against a posed RGB image with corresponding object mask (obtained, for example, with off-the-shelf segmentation networks).
\cref{fig:scannet_rec} shows qualitative results for this single-image reconstruction task on chairs from ScanNet~\cite{dai2017scannet}. 
Furthermore, we show results of our model conditioned on CLIP-embeddings~\cite{radford2021learning} in the appendix.

\subsection{Limitations}
While our method shows promising results on the tasks of conditional and unconditional radiance field synthesis, several limitations remain.
Compared to GAN-based approaches, our radiance fields-based approach requires a sufficient number of posed views in order to generate good training samples and suffers from lower sampling times.
In this context, it would be interesting to explore leveraging faster sampling methods~\cite{kong2021fast}.
Finally, our model is constrained in the maximum grid resolution by training-time memory limitations.
These could be addressed by exploring adaptive ~\cite{takikawa2021nglod, Wang-2017-ocnn} or sparse grid structures~\cite{fridovich2022plenoxels,takikawa2021nglod, schwarzvoxgraf} as well as factorized neural fields representations \cite{chen2022tensorf, wang2022rodin}.
\section{Conclusions}
We introduce \Ours{} -- a novel approach for 3D radiance field synthesis based on denoising diffusion probabilistic models.
To the best of our knowledge, \Ours{} is the first generative diffusion-based method to operate directly on volumetric radiance fields.
Our model learns multi-view consistent priors from collections of posed images, enabling free-view image synthesis and accurate shape generation.
We evaluated \Ours{} on several object classes, comparing its performance against state-of-the-art GAN-based approaches, and demonstrating its effectiveness in both conditional and unconditional 3D generation tasks.

\section*{Acknowledgements}
This work was done during Norman’s and Yawar’s internships at Meta Reality Labs Zurich as well as at TUM, funded by a Meta SRA.
Matthias Nie{\ss}ner was also supported by the ERC Starting Grant Scan2CAD (804724).

{\small
\bibliographystyle{ieee_fullname}
\bibliography{main}
}

\newpage
\appendix

\newpage
\section*{\Large\textbf{Appendix}}

In this supplementary document, we discuss additional details about our method, the data used for training and evaluation, and show further qualitative results. We also refer to our  for a comprehensive overview with further qualitative results. 

\begin{figure*}
\begin{center}
\includegraphics[width=1.80\columnwidth, trim={0cm 11cm 0 0cm},clip, page=2]{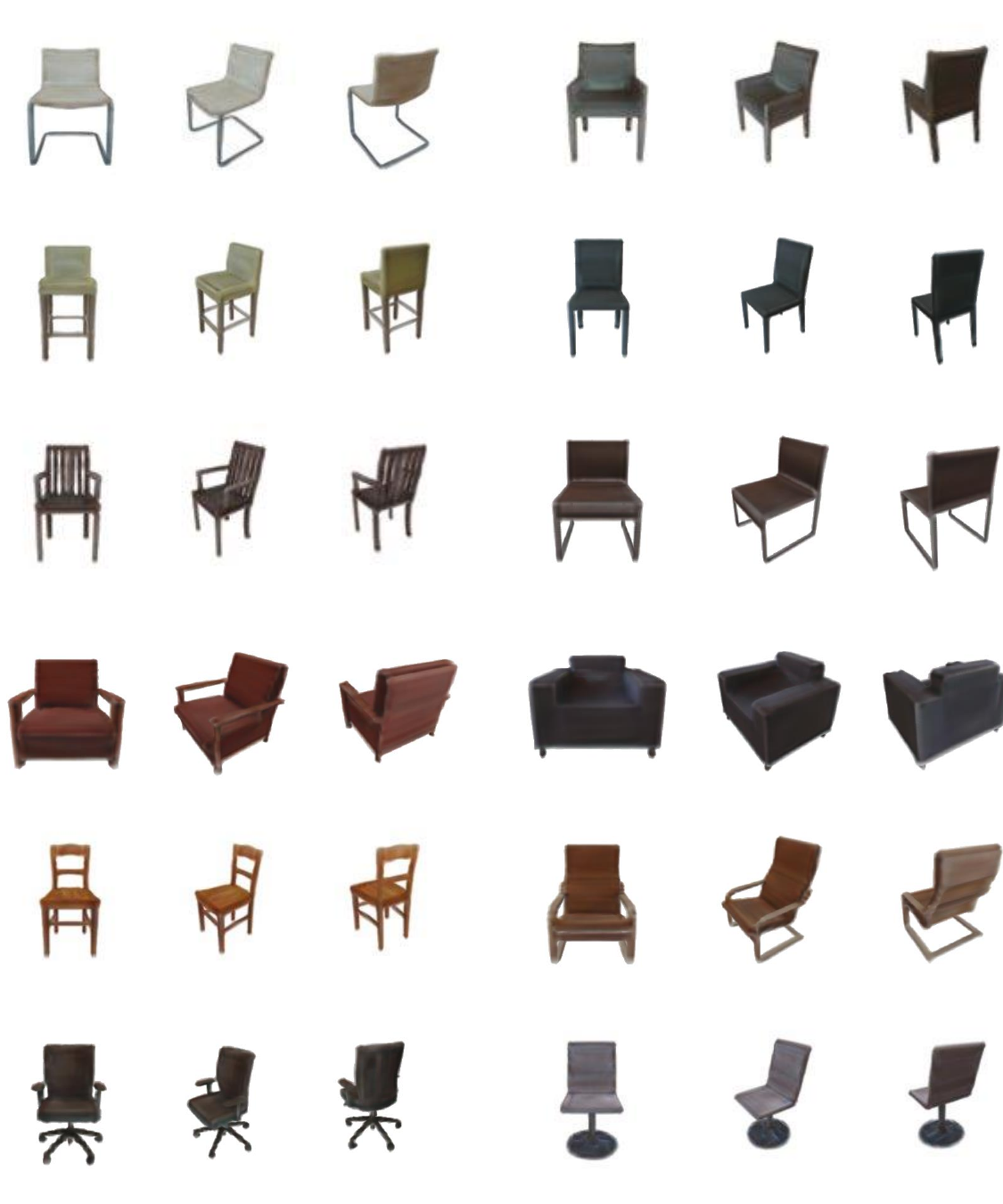}

\end{center}
\vspace{-1em}
\caption{
Additional qualitative sampling results on PhotoShape Chairs~\cite{photoshape2018}.
}
\label{fig:add_chairs0}

\end{figure*}

\begin{figure*}
\begin{center}
\includegraphics[width=1.85\columnwidth, trim={0cm 14cm 0cm 0.0cm},clip, page=4]{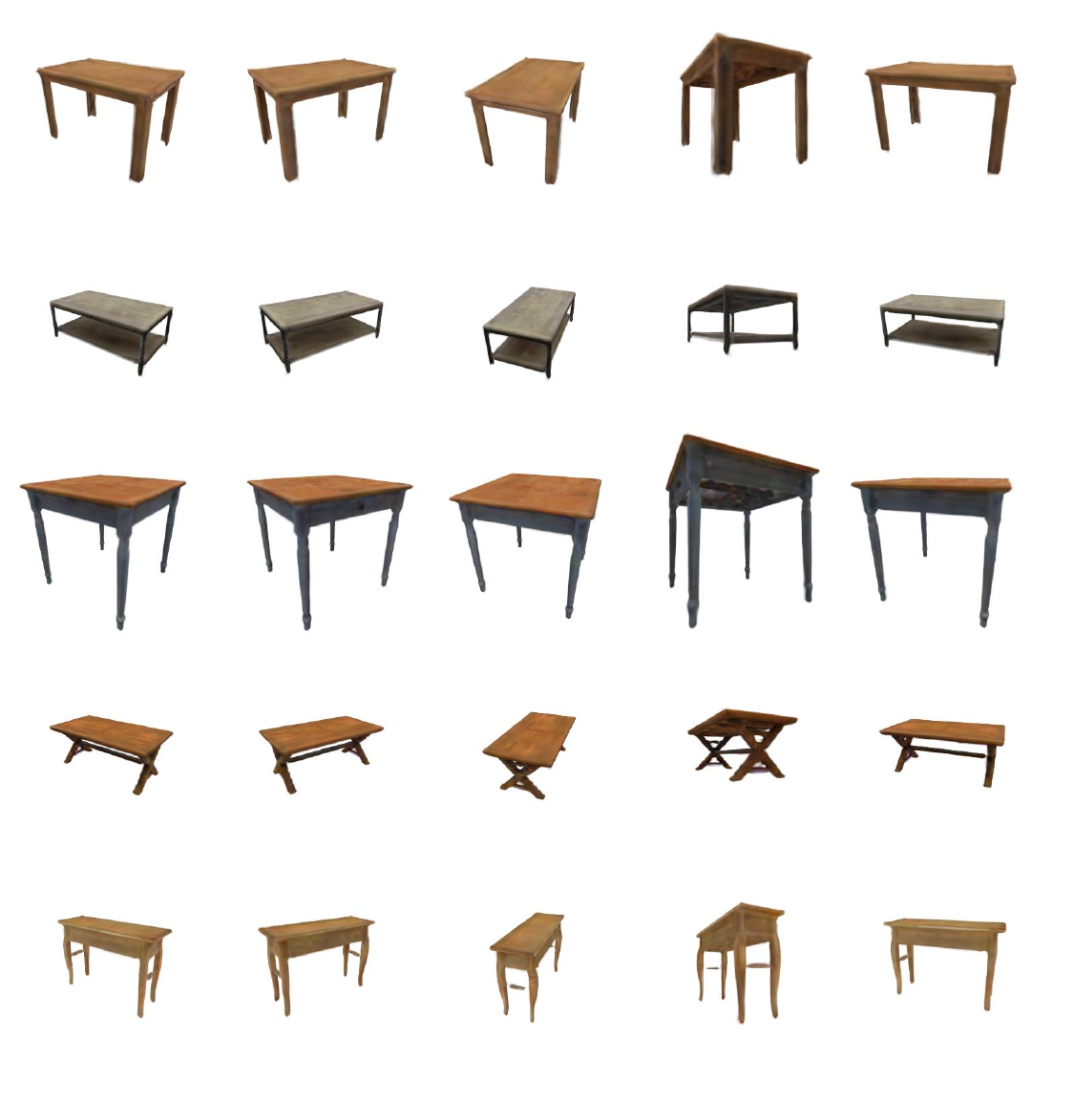}

\end{center}
\vspace{-2.2em}
\caption{
Additional qualitative sampling results on ABO Tables~\cite{collins2022abo}.
}
\label{fig:add_tables0}

\end{figure*}

\begin{figure*}
\begin{center}
\includegraphics[width=1.80\columnwidth, trim={0cm 11cm 0 0cm},clip, page=1]{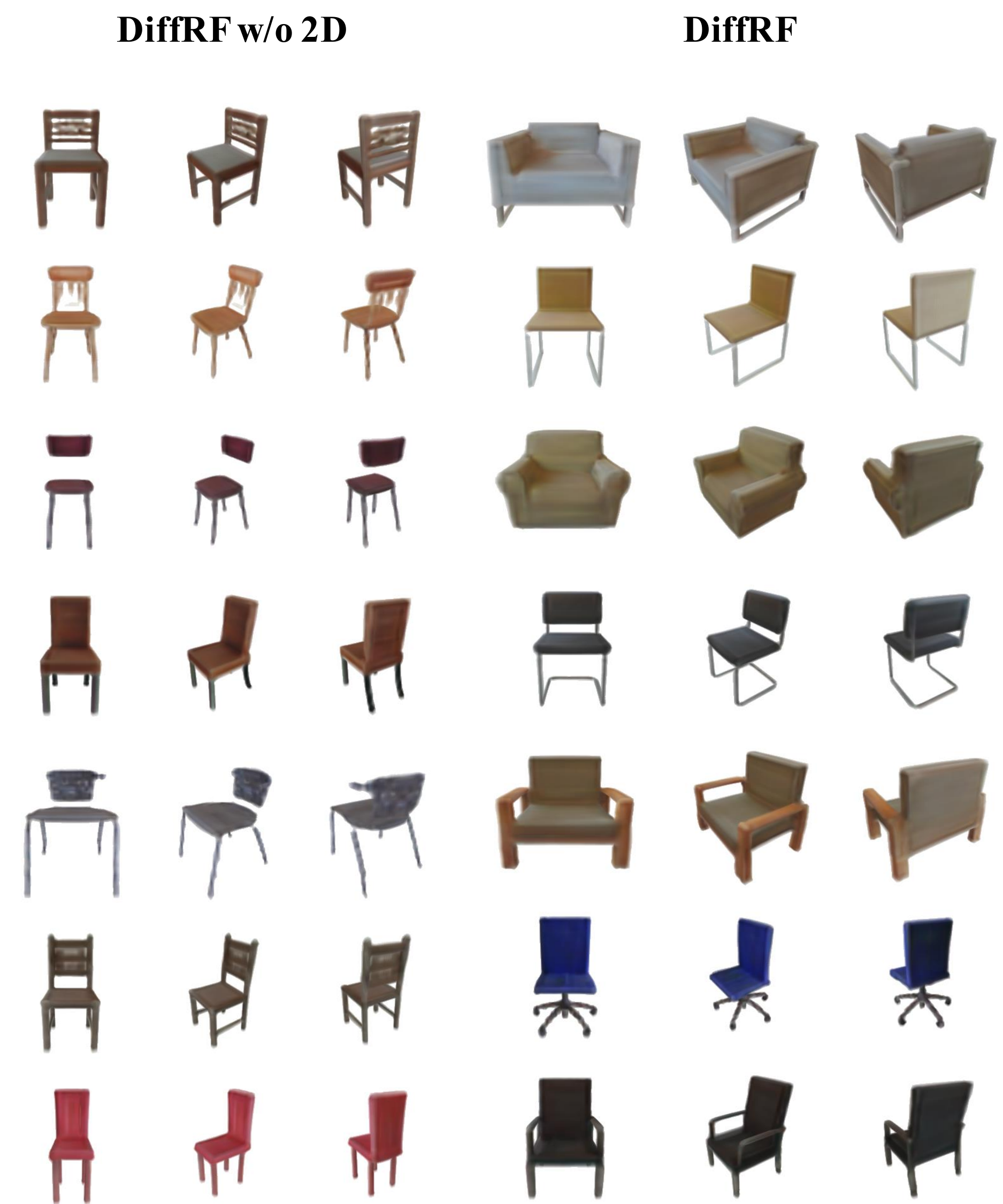}

\end{center}
\vspace{-1em}
\caption{
Qualitative comparison on PhotoShape Chairs~\cite{photoshape2018} when removing the 2D rendering loss.
}
\label{fig:comp_2d_ps}

\end{figure*}

\begin{figure*}
\begin{center}
\includegraphics[width=1.80\columnwidth, trim={0cm 11cm 0 0cm},clip, page=3]{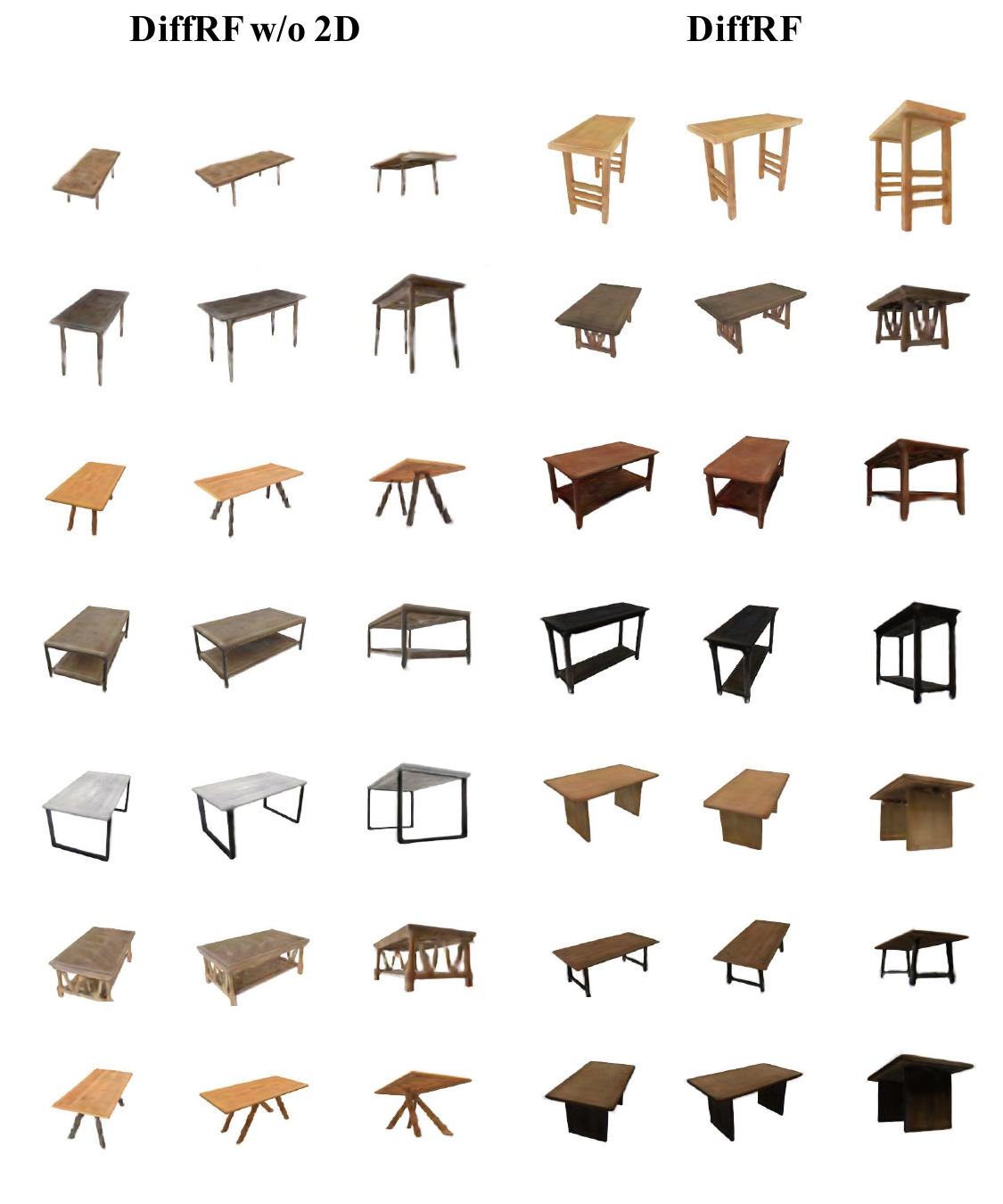}

\end{center}
\vspace{-1em}
\caption{
Qualitative comparison on  ABO Tables~\cite{collins2022abo} when removing the 2D rendering loss.
}
\label{fig:comp_2d_abo}

\end{figure*}

\section{Additional qualitative results}

We provide additional qualitative results on PhotoShape Chairs~\cite{photoshape2018} in \cref{fig:add_chairs0} as well as on ABO Tables~\cite{collins2022abo} in \cref{fig:add_tables0}. Furthermore, we provide a qualitative comparison of our method when removing the rendering loss in \cref{fig:comp_2d_ps} and \cref{fig:comp_2d_abo}.

\section{Implementation detail}
\paragraph{Architecture}
We base our architecture on 2D U-Net structure from~\cite{dhariwal2021diffusion}: For this, we replace the 2D convolutions in the ResNet and attention blocks with corresponding 3D convolutions, preserving kernel sizes and strides. Furthermore, we use 3D average pooling layers instead of 2D in the downsampling steps. 
Our U-Net consists of $4$ scaling blocks with two ResNet blocks per scale, where we linearly increase the initial feature channel dimension of $64$ to $256$. We use skip attention blocks at the scaling factors $2$, $4$, and $8$ with $32$ channels per head. 

\paragraph{Training details}

We train all models with a batch size of 8 and use the Adam optimizer with an initial learning of $10^{-4}$. We apply a linear beta scheduling from $0.0015$ to $0.05$ at $1000$ timesteps. From $4$ random training views at a resolution of $128\times 128$, we sample $8192$ random pixels for the rendering supervision (with $92$ z-steps for volumetric rendering) and weight the rendering loss with $\omega_t=\bar\alpha_t^2$.
We train for $3.0$m iterations with a decaying LR scheduling for $10^{-4}$ to $10^{-6}$ at a voxel grid resolution of $32$ on $2$ GPUs on every data set. 

\paragraph{Sampling time}
We perform DDPM sampling for $1000$ iterations leading to a run time of $48.6$s per sample on an NVIDIA RTX 2080 TI. Once synthesized, our explicit representation enables rendering at $128\times128$ resolution with over 380 FPS.
\section{Data}

\paragraph{Radiance Field Generation}
For PhotoShape Chairs, we render the provided 15,576 chairs using Blender Cycles from 200 views on an Archimedean spiral at a fixed radius of $2.5$ units with pitch starting from $-20 ^{\circ}$ to $60^{\circ}$. For ABO Tables, we use the provided 91 renderings with 2-3 different environment map settings per object, resulting in 1676 tables. 
For PhotoShape Chairs, we hold out $10\%$ of the samples for testing based on shape ids selected randomly, whereas for ABO Tables, we use the official data split. We fit explicit voxel grids at a resolution of $32^3$ using volumetric rendering with spherical harmonics of degree $2$ for an initial fit. We then fine-tune our representations for spherical harmonics of degree $0$, which we found to lead to sharper geometry compared to directly optimizing density and color features. 
We furthermore bound the feature space to $[-1,1]$ which we found to stabilize the sampling process noticeably affecting the rendering quality.

\paragraph{Evaluation}
For image quality evaluation, we calculate FID and IS by sampling 10k views by rendering $1000$ samples from $10$ random views at a resolution of $128\times128$. 
We follow ~\cite{pointflow} and evaluate the geometric quality by computing the Coverage Score (COV) and Minimum Matching Distance (MMD) using Chamfer Distance (CD)
\begin{align*}
    \textrm{CD}(X,Y)&= \sum_{x\in X}\min_{y\in Y}||x-y||^2_2 + \sum_{y\in Y}\min_{x\in X}||x-y||^2_2, \\
    \textrm{COV}(S_g,S_r)&=\frac{|\{\textrm{arg min}_{Y\in S_r} CD(X,Y)|X\in S_g\}|}{|S_r|},\\
    \textrm{MMD}(S_g,S_r)&=\frac{1}{|S_r|}\sum_{Y\in S_r}\min_{X\in S_g} CD(X,Y),
\end{align*}
on a reference set $S_r$ (the test samples) and a generated set $S_g$ twice as large as the reference set. 
We extract meshes using marching cubes~\cite{lorensen1987marching} and sample $2048$ points on the faces. To account for potentially different scaling of the samples produced by the 3D-aware GAN models, we normalize all point clouds by centering in the origin and an-isotropic scaling of the extent to $[-1,1]$. 

For evaluation of the masked radiance field completion, we additionally compute a masked peak signal-to-noise ratio (mPSNR): Given a binary mask $m$ of the input radiance field $f^{in}$, we compute for each corresponding input image the non-masked area by depth-based projection into the image plane using depth estimated from the input radiance field. We then compute the mPSNR by averaging the PSNR on the non-masked pixels for all evaluation views (we choose 10 views randomly).

\section{Conditional sampling}
\paragraph{Masked completion}
Since the generator of EG3D~\cite{Chan2022} is trained via 2D discriminator guidance, we perform 3D masked completion via GAN inversion. For this, we start from a random initial latent code and repeat the following steps for 200 iterations on each masked sample: We render the current synthesized sample from $8$ views and project the 3D input mask onto the synthesized views using the predicted depths. On the remaining non-masked regions, we compute the photometric error with the input images. We use the Adam optimizer with a learning rate of $10^{-2}$ with a small $L_2$ regularization term on the code (weighted with $5\times10^{-2}$) in order to update the latent code. 

\paragraph{Image-to-Volume Synthesis}
Given a posed and segmented image, we condition our trained radiance field diffusion model by steering the sampling processing similar to the Classifier Guidance formulation from~\cite{dhariwal2021diffusion}: 
During sampling time, for each time step $t$, we gradually update the predicted denoised field 
$\tilde f^t_0$ towards minimizing the photometric error obtained from comparing the rendering $\tilde{I}_t$ from a given pose with the foreground-masked target image $I$. For this, we compute the gradient $\nabla_{\tilde f^t_0}(\tilde{I_t}, I )$ on the current denoising estimate by volumetric rendering and steer the sampling process by $\tilde f^t_0\leftarrow \tilde f^t_0 - \lambda\nabla_{\tilde f^t_0}(\tilde{I_t}, I)$ with a small guidance weight $\lambda$.  

\section{CLIP conditioning}
\begin{figure*}
\begin{center}
\includegraphics[width=1.65\columnwidth, trim={0cm 12cm 0 0cm},clip, page=8]{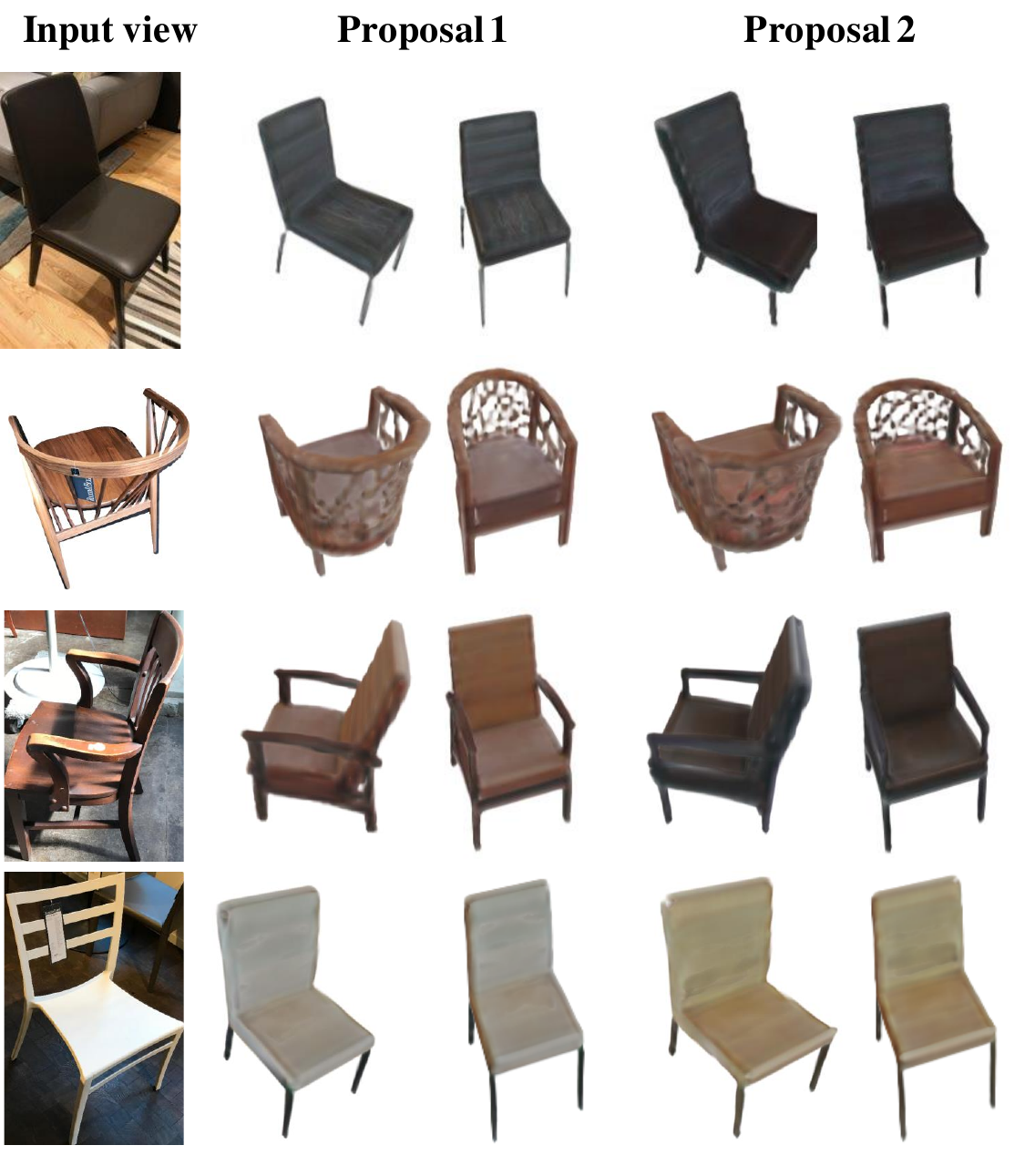}

\end{center}
\vspace{-1em}
\caption{
Image-conditional inference using CLIP-embeddings on Pix3D~\cite{sun2018pix3d} images.
}
\label{fig:pix3d_cond}

\end{figure*}

\begin{figure*}
\begin{center}
\includegraphics[width=1.85\columnwidth, trim={0cm 9cm 0 0cm},clip, page=7]{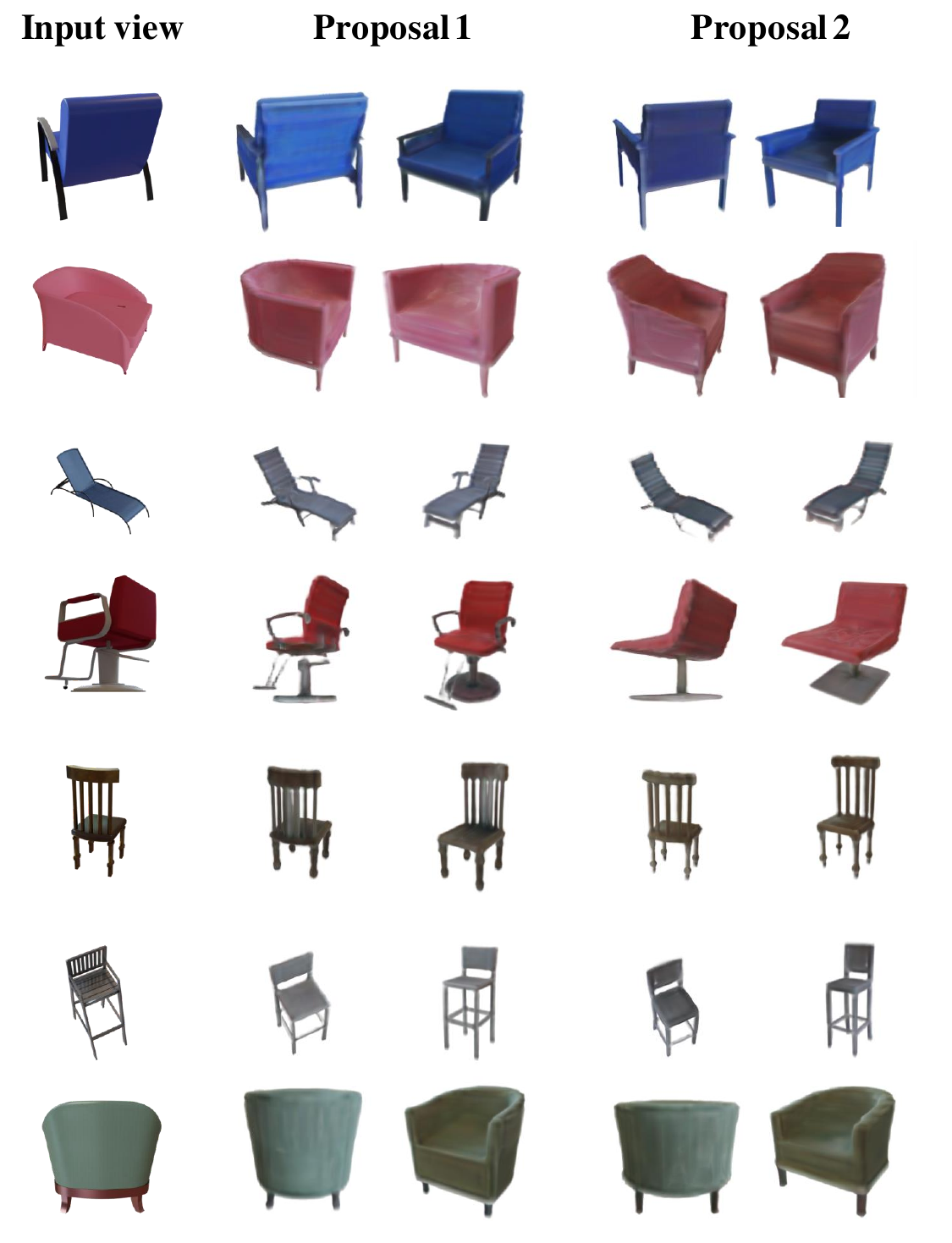}

\end{center}
\vspace{-1em}
\caption{
Image-conditional inference using single-view CLIP-embeddings on PhotoShape Chairs~\cite{photoshape2018}.
}
\label{fig:img_cond}

\end{figure*}

Following related work~\cite{cheng2022sdfusion, zeng2022lion, wang2022rodin}, we additionally augment our model to condition on embeddings derived from text or single-image encodings obtained from CLIP ViT-B/32 ~\cite{radford2021learning} using cross-attention layers. For training, we use random single training views encoded by the frozen CLIP model to condition the denoiser. Here, we adapt the cross-attention mechanism from~\cite{dhariwal2021diffusion} for the 3D U-Net and do not train the image encoder in order to preserve the image-text-correspondence of CLIP. 
We show examples on single-image PhotoShape samples in \cref{fig:img_cond} as well as on real-world image from the Pix3D dataset~\cite{sun2018pix3d} in \cref{fig:pix3d_cond}. 

\begin{figure*}
\begin{center}
\includegraphics[width=1.85\columnwidth, trim={0cm 18cm 0 0cm},clip, page=6]{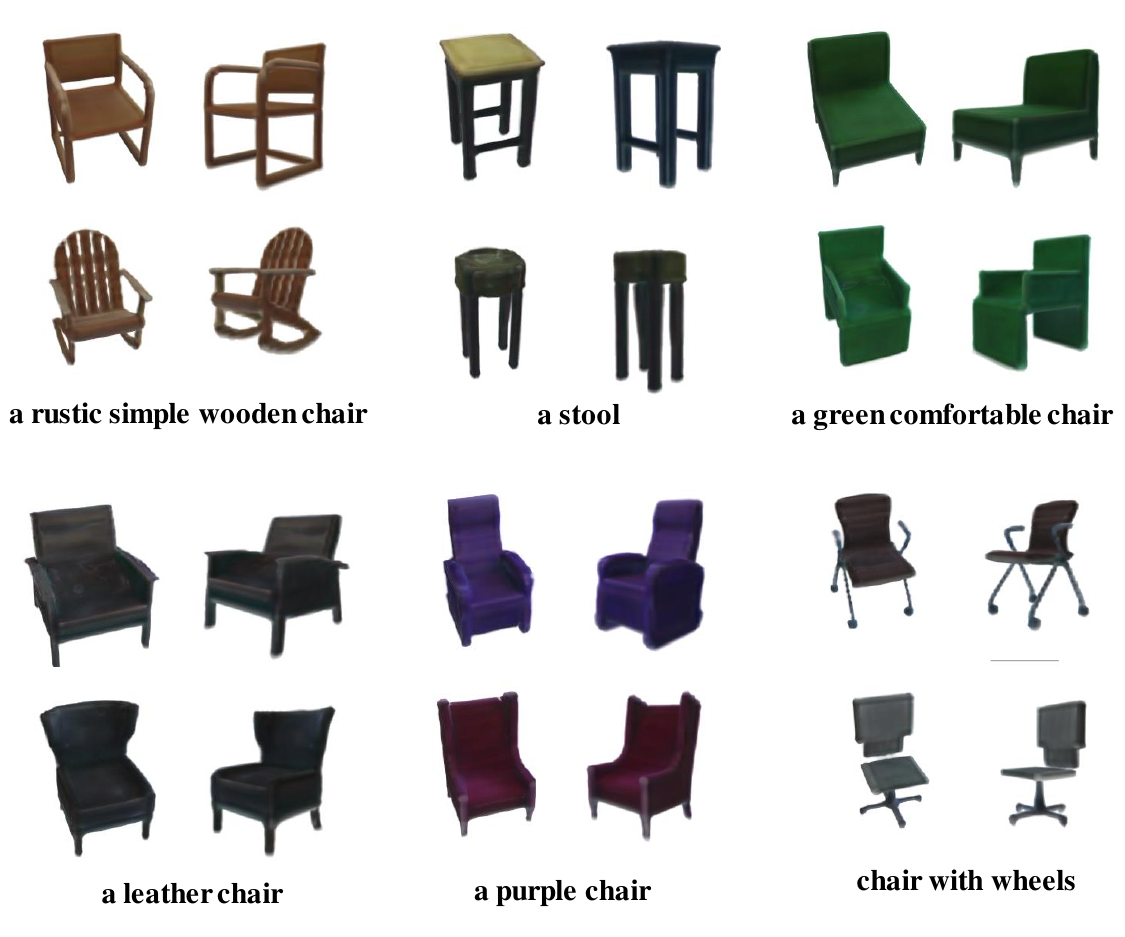}

\end{center}
\vspace{-1em}
\caption{
Text-conditional inference using CLIP-embeddings trained on PhotoShape Chairs ~\cite{photoshape2018}.
}
\label{fig:text_cond}

\end{figure*}

As these codes have strong correspondences to text samples by design, we can guide the sampling process by text prompts, examples are shown in \cref{fig:text_cond} without the need for training on text-radiance field pairs.

\end{document}